\documentclass[runningheads]{llncs}
\usepackage[T1]{fontenc}
\usepackage{graphicx}
\usepackage{amsmath}
\usepackage{amssymb}
\usepackage{booktabs}
\usepackage{mfirstuc}

\usepackage{geometry}
\usepackage{changepage}
\usepackage{float}
\usepackage{multirow}
\usepackage{cite}
\usepackage{algorithm}
\usepackage{algpseudocode}

\usepackage[hidelinks]{hyperref}
\begin{document}
\title{Advancing Question Answering on Handwritten Documents: A State-of-the-Art Recognition-Based Model for HW-SQuAD}
%
%
\author{Aniket Pal\orcidID{0000-0002-9971-8674} \and
Ajoy Mondal\orcidID{0000-0002-4808-8860} \and
C.V. Jawahar\orcidID{0000-0001-6767-7057}}
\authorrunning{Aniket et al.}
%
\institute{CVIT, IIIT Hyderabad
\email{aniket.pal@research.iiit.ac.in}, \email{ajoy.mondal, jawahar@iiit.ac.in}\\
 }
\maketitle              
\begin{abstract}
Question-answering on handwritten documents is a challenging task with numerous real-world applications. This paper proposes a novel recognition-based approach that improves upon the previous state-of-the-art on the HW-SQuAD and BenthamQA datasets. Our model incorporates transformer-based document retrieval and ensemble methods at the model level, achieving an Exact Match score of 82.02\% and 69\% in HW-SQuAD and BenthamQA datasets, respectively, surpassing the previous best recognition-based approach by 10.89\% and 3\%. We also enhance the document retrieval component, boosting the top-5 retrieval accuracy from 90\% to 95.30\%. Our results demonstrate the significance of our proposed approach in advancing question answering on handwritten documents. The code and trained models will be publicly available to facilitate future research in this critical area of natural language. 

\end{abstract}
\section{Introduction}
Question Answering \cite{rajpurkar-etal-2016-squad, SQuAD_2} has become an important task in NLP that aims to automatically provide correct answers to questions articulated in natural language processing. The ability to answer questions effectively with respect to a given context is widely useful, including but not limited to information retrieval, knowledge management, and intelligent personal assistants. The Stanford Question Answering Dataset (SQuAD) \cite{rajpurkar-etal-2016-squad} has become a widely-used benchmark for evaluating the performance of QA systems on text-based documents, which consists of a collection of Wikipedia articles and crowd-sourced questions, where the answers are spans of text extracted from the corresponding articles.

However, answering questions on handwritten document images introduces additional challenges compared to traditional text-based QA. Handwritten documents exhibit complex layouts, varying writing styles, and noise and distortions, making recognizing and extracting relevant information challenging. Furthermore, handwritten text recognition is challenging, requiring robust models that can handle the variability and ambiguity inherent in handwritten content.

To address these challenges, the HW-SQuAD dataset \cite{HW-SQUaD} was introduced as an extension of SQuAD to the handwritten domain. It consists of synthetic handwritten document images paired with questions and answers, where the answers are spans of text within the documents. This dataset has spurred research into developing QA systems that can effectively handle the complexities of handwritten documents. The two main approaches for tackling HW-SQuAD are recognition-based methods, which rely on accurate handwritten text recognition to convert the images into machine-readable text and apply traditional text-based QA techniques, and recognition-free methods, which directly process the handwritten images without explicit text recognition, leveraging visual features and spatial layout information to locate the answer spans.

Previous works on HW-SQuAD have explored both recognition-based and recognition-free approaches. Minesh et al. \cite{HW-SQUaD} proposed a recognition-based method that combines handwritten text recognition with a pre-trained language model for question answering. Their approach achieved an Exact Match score of 70\% on the HW-SQuAD dataset. In the recognition-free domain, the same author introduced a visual QA model that directly operates on the handwritten document images, achieving an accuracy (snippet extraction accuracy \cite{HW-SQUaD}) of 15.9 \%. Despite these advancements, there remains significant room for improvement in QA performance on handwritten documents.

The previous recognition-based model consists of two main components: the document retriever and the document reader. The document retriever employs the naive TF-IDF algorithm to rank and select relevant documents based on the question. In contrast, the document reader utilizes the BERT QA \cite{BERT} model to extract the answer span from the retrieved documents. 

In this paper, we propose a novel document retriever that combines the strengths of the TF-IDF algorithm and sentence transformers, resulting in significant improvements in document retrieval performance compared to the previous state-of-the-art. Furthermore, we introduce advanced pre-processing techniques that further enhance the accuracy of the retrieval process. We employ an ensemble approach for the document reader component that leverages the BERT QA model and other extractive QA models such as SpanBERT and DeBERTa, enabling more robust and accurate answer extraction from the retrieved documents.

The main contributions of our work are as follows:
\begin{itemize}
    \item We propose a novel document retriever that effectively combines the TF-IDF algorithm with sentence transformers, significantly improving the retrieval performance on the HW-SQuAD dataset.
    \item We introduce advanced pre-processing techniques that further enhance the accuracy of the document retrieval process, enabling more precise selection of relevant documents for the given question.
    \item We employ an ensemble approach for the document reader component, leveraging the strengths of multiple extractive QA models, including BERT and DeBERTa, to achieve more robust and accurate answer extraction.
    \item Through extensive experiments, we demonstrate the superiority of our proposed approach, surpassing the previous state-of-the-art performance on the HW-SQuAD and BenthamQA datasets.
\end{itemize}

The remainder of this paper is organized as follows. Section 2 discusses related work on question answering and handwritten document processing. Section 3 describes our proposed recognition-based approach in detail, including the handwritten text recognition model, question-document matching mechanism, and enhanced retrieval method. Section 4 presents the experimental setup, dataset details, and evaluation metrics. Section 5 reports and analyzes the results of our experiments, comparing our approach with previous state-of-the-art methods. Section 6 provides a discussion on the implications of our findings, the limitations of our approach, and potential future research directions. Finally, Section 7 concludes the paper and summarizes our contributions.

\section{Related Works}

The natural language processing (NLP) and information retrieval (IR) communities have been actively researching machine reading comprehension (MRC) and open domain question answering (QA). The introduction of large-scale datasets like SQuAD \cite{rajpurkar-etal-2016-squad}, MS MACRO \cite{MS_MACRO}, and Natural Questions \cite{Natural_Questions} has spurred the development of deep learning-based QA/MRC systems \cite{BERT}\cite{28_link}-\cite{30_link} capable of answering questions about a given text corpus or passage. Our work builds upon these advancements but focuses on answering questions on handwritten document images using recognition-based approaches.
Visual question answering (VQA) has gained significant attention in the computer vision community in recent years [10,31–35]. Early VQA datasets and methods often ignored text in images, treating the problem as multi-class classification with a fixed set of answers and emphasizing visual aspects like objects and attributes. However, Gurari et al. \cite{36_link} showed that answering many questions asked by visually impaired individuals on their own images necessitates reading and interpreting the image text. This led to the creation of Scene Text VQA \cite{scene_text} and TextVQA \cite{Text-book-QA} datasets, where reading image text is essential for answering questions. Our work differs from these tasks in two main aspects: (1) we focus on handwritten document images instead of "in the wild" images with scattered text tokens, and (2) we aim to answer questions on a collection of document images rather than a single image.
Other relevant VQA works include VQA on charts and plots \cite{DVQA}\cite{Figure_QA} and Textbook Question Answering (TQA) \cite{Text-book-QA}. TQA seeks to answer questions given a context of text, diagrams, and images, but the textual information is provided in computer-readable format. For VQA on charts and plots, OCR is required to recognize the text and answer many questions. However, the text in these synthetically generated charts/plots is sparse and rendered in standard fonts, unlike the handwritten sentences and paragraphs in our case.
Our work is inspired by the DocVQA dataset \cite{DocVQA}, which includes a wide variety of document images containing printed, typewritten, handwritten, and born-digital text in the form of sentences, forms, tables, figures, and photographs. While DocVQA Task 1 adheres to the standard VQA setting with textual answers, we propose an improved recognition-based approach to answer questions on handwritten document collections like HW-SQuAD and BenthamQA.
In the field of information retrieval and keyword spotting, there have been numerous efforts on handwritten document indexing and retrieval \cite{40_link}\cite{44_link}. The ImageCLEF 2016 Handwritten Scanned Document Retrieval challenge \cite{44_link} aimed to develop retrieval systems for handwritten documents. Although there are similarities between this challenge and our work, such as using multi-token queries and retrieving document segments or snippets, the document retrieval task differs in two aspects: (1) queries are search queries, not natural language questions, and (2) the task requires all query tokens to appear in the same order in the retrieved snippet.
Kise et al. \cite{45_link} tackled document retrieval for building a QA system on a collection of printed document images, which is likely the first work on QA over a document image collection. They used documents with machine-printed English text, which is relatively easier to recognize compared to handwritten text. Our work advances this line of research by proposing an improved recognition-based approach for QA on handwritten document collections like HW-SQuAD and BenthamQA.
DocVQA Task 2 \cite{DocVQA} is similar to our work in that both deal with QA over a document collection. However, DocVQA Task 2 uses a collection of US candidate registration forms with the same template, while we focus on handwritten documents with diverse content. Moreover, DocVQA Task 2 aims to retrieve all documents required to answer the question correctly, while our approach focuses on returning precise answer snippets.

In this work, we improved the recognition-based approach in \cite{HW-SQUaD} for Question Answering on handwritten document collection. We have improved the Document Retriever by adding advanced preprocessing and the Transformer. In addition, we have improved the Document Reader part by using the Ensemble method. Our proposed end-to-end recognition-based pipeline achieved state-of-art in the HW-SQuAD and Bentham-QA datasets.

\begin{figure*}[!ht]
\hspace{-0.2cm}
    \includegraphics[width=\textwidth]{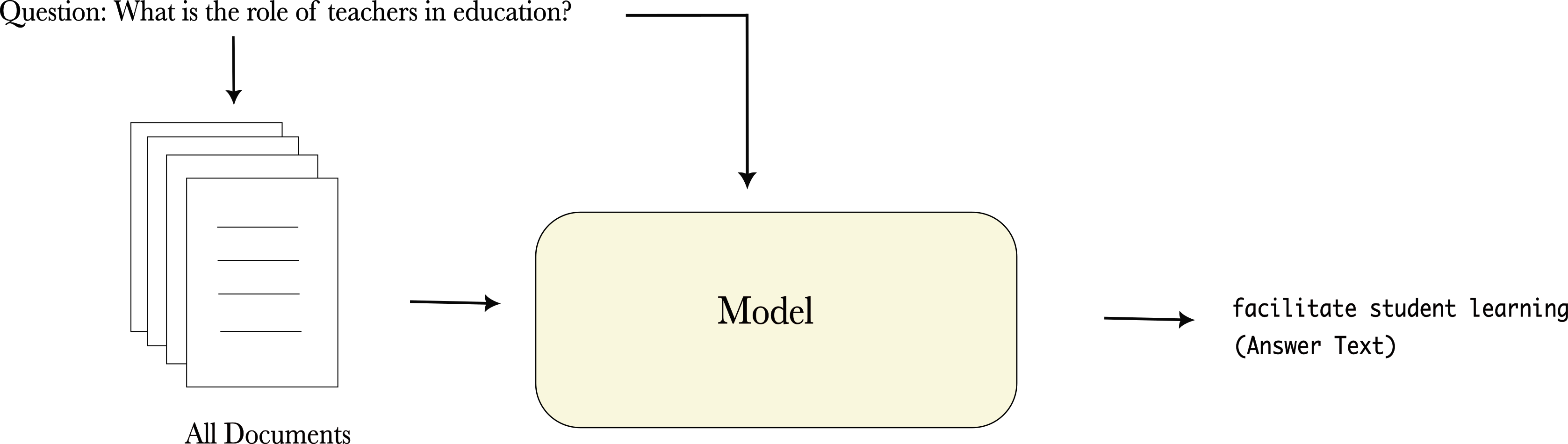}
    \caption{Overview of our problem statement. The Question along with all th documents are fed to the model and it needs to predict the answer.}
    \label{fig:enter-label}
\end{figure*}

\section{Method}

This section will delineate the methodology of our proposed end-to-end pipeline for Question Answering. Our problem is illustrated in Figure 1. The model is provided with the question and all documents to predict the answer. The model may be either recognition-based or recognition-free. Here, the Question ``What is the role of teachers in education?'' is fed to the model along with all the documents, and the model needs to predict the answer ``facilitate student learning.'' The model may be either recognition-based or recognition-free. 

We have followed the recognition-based architecture \cite{HW-SQUaD}. This architecture is comprised of two parts: i) Document Retriever and ii) Document Reader part. The Document Retriever part comprises TF-IDF and preprocessing, which retrieve the top k document from all the document collections. For the Document Reader part, The BERT Large model is used.

In our proposed architecture, we improve the Document Retriever as well as the Document Reader parts. For document retrieval, we use additional preprocessing techniques, such as the Sentence transformer and TF-IDF. We used an ensemble of two large BERTs and one DeBERTa model for the Document Reader part. Our proposed architecture significantly improves document retrieval and reader performance. Figure 3 depicts the entire architecture of our proposed model.

This section is divided into two parts: first, we discuss our proposed Document Retriever, and then we discuss the Document Reader.

\begin{figure*}[ht]
\hspace{-0.2cm}
    \includegraphics[width=\textwidth]{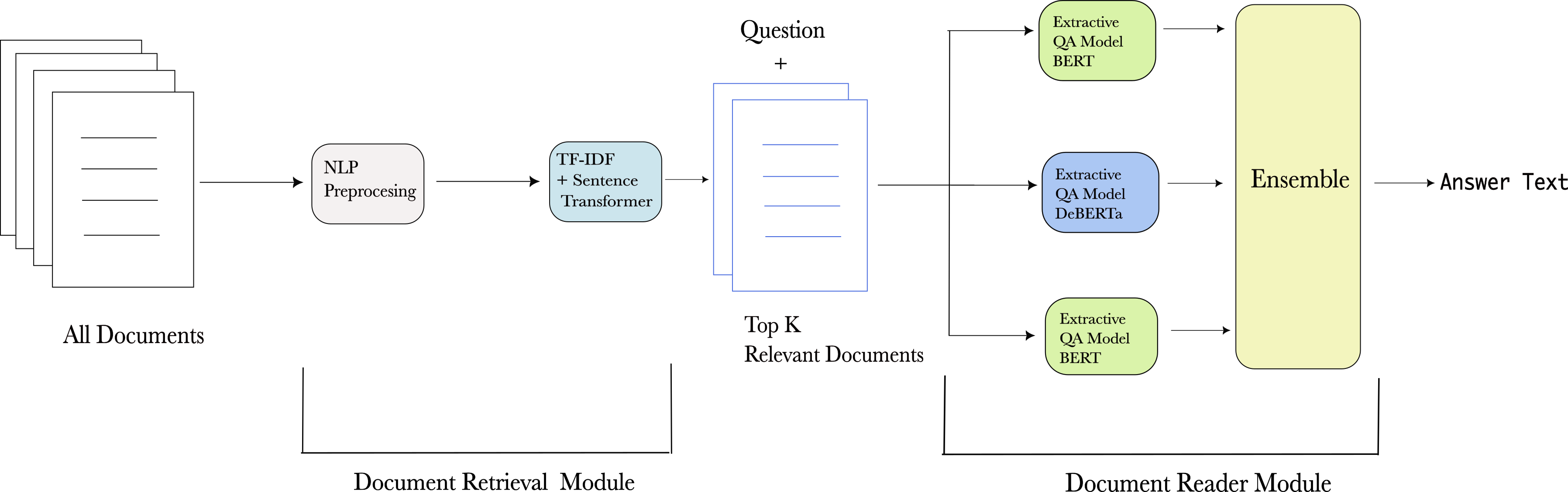}
    \caption{Entire workflow and architecture of Our propose Recognition Approach. The Document Retrieval module is consist of NLP pre-processing, TF-IDF and Sentence transformer. In the Document Reader module we implement ensemble technique with two BERT large and one DeBERTa Large model.}
    \label{fig:enter-label}
\end{figure*}

\subsection{Document Retriever Module}

In the document retriever component of our approach, we propose a novel technique that enhances the traditional TF-IDF algorithm by incorporating sentence transformers and advanced NLP preprocessing steps. TF-IDF is a well-established method in information retrieval that assigns weights to terms based on their frequency within a document and their rarity across the corpus. Let $V$ be the vocabulary of unique terms across all documents in $D_{\text{processed}}$. The TF-IDF vectorization can be represented as follows:

\subsection*{Vectorization}

2. \textbf{Vectorizer Transformation:}

For a set of contexts \( C = \{c_1, c_2, \ldots, c_n\} \), where each context \( c_i \) is preprocessed:

\begin{itemize}
    \item \textbf{TF-IDF Vectorization:} The TF-IDF (Term Frequency-Inverse Document Frequency) vectorization process converts a collection of text documents into a matrix representation, highlighting the importance of terms in each document. For a set of contexts \( C = \{c_1, c_2, \ldots, c_n\} \):

a) \textbf{Term Frequency (TF):}

   The term frequency \( \text{TF}(t, d) \) of term \( t \) in document \( d \) is defined as:

   \[
   \text{TF}(t, d) = \frac{f_{t,d}}{ \sum_{t' \in d} f_{t',d} }
   \]

   where \( f_{t,d} \) is the raw count of term \( t \) in document \( d \), and the denominator is the total count of all terms in the document \( d \).

b) \textbf{Inverse Document Frequency (IDF):}

   The inverse document frequency \( \text{IDF}(t, D) \) of term \( t \) across a corpus \( D \) (collection of all documents) is defined as:

   \[
   \text{IDF}(t, D) = \log \left( \frac{N}{1 + |\{d \in D : t \in d\}|} \right)
   \]

   where \( N \) is the total number of documents in the corpus \( D \), and \( |\{d \in D : t \in d\}| \) is the number of documents containing the term \( t \). The term \( 1 \) in the denominator is added to prevent division by zero.

c) \textbf{TF-IDF Calculation:}

   The TF-IDF score for term \( t \) in document \( d \) is given by:

   \[
   \text{TF-IDF}(t, d, D) = \text{TF}(t, d) \times \text{IDF}(t, D)
   \]

d) \textbf{Document-Term Matrix:}

   For the set of contexts \( C = \{c_1, c_2, \ldots, c_n\} \), the TF-IDF vectorization can be represented as a document-term matrix \( \mathbf{M}_{\text{TF-IDF}} \), where each entry \( \mathbf{M}_{\text{TF-IDF}}[i, j] \) represents the TF-IDF score of term \( t_j \) in document \( c_i \):

   \[
   \mathbf{M}_{\text{TF-IDF}}[i, j] = \text{TF-IDF}(t_j, c_i, C)
   \]

Thus, the TF-IDF vectorization process for the set of contexts \( C \) is represented as:

\[
\mathbf{M}_{\text{TF-IDF}} = \text{TF-IDF Vectorizer}(C)
\]

where \( \mathbf{M}_{\text{TF-IDF}} \) is the resulting document-term matrix.

\vspace{0.2cm}

\end{itemize}

3. \textbf{Transformer Encoding:}

The Sentence Transformer model converts a collection of text documents into a matrix of dense vector representations. For a set of contexts \( C = \{c_1, c_2, \ldots, c_n\} \):

a) \textbf{Context Embedding:}

   Each context \( c_i \in C \) is passed through the Sentence Transformer model to obtain its dense vector representation (embedding). Let \( \text{ST}(\cdot) \) denote the transformation function of the Sentence Transformer model. The embedding \( \mathbf{e}_{c_i} \) for context \( c_i \) is given by:
 
   \[
   \mathbf{e}_{c_i} = \text{ST}(c_i)
   \]

   where \( \mathbf{e}_{c_i} \in \mathbb{R}^d \) and \( d \) is the dimensionality of the embedding space.

b) \textbf{Embedding Matrix:}

   For the entire set of contexts \( C \), the Sentence Transformer model produces a matrix of embeddings. This matrix \( \mathbf{E}_C \) is constructed by stacking the embeddings of all contexts:

   \[
   \mathbf{E}_C = 
   \begin{bmatrix}
   \mathbf{e}_{c_1} \\
   \mathbf{e}_{c_2} \\
   \vdots \\
   \mathbf{e}_{c_n}
   \end{bmatrix}
   \]

   where \( \mathbf{E}_C \in \mathbb{R}^{n \times d} \) is the matrix of encoded vectors, with each row representing the embedding of a context.

Thus, the encoding process using the Sentence Transformer model for the set of contexts \( C \) is represented as:

\[
\mathbf{E}_C = \text{SentenceTransformer}(\text{model}, C)
\]

where \( \mathbf{E}_C \) is the resulting matrix of encoded vectors obtained from the Sentence Transformer model.

\begin{algorithm}
\caption{Steps for Document Retrieval}\label{alg:cap}
\begin{algorithmic}[1]
\State \textbf{Query Vectorization and Encoding}

\textbullet~ \textbf{Input:} Query \( q \)

\textbullet~ Preprocess the query:
\[
q' = \text{preprocess}(q)
\]

 \textbullet~ Vectorize using the same vectorizers:
\[
\mathbf{q}_{\text{TF-IDF}} = \text{TF-IDF Vectorizer.transform}(q')
\]

 \textbullet~ Encode using the transformer model:
\[
\mathbf{e}_q = \text{SentenceTransformer.encode}(\text{model}, q')
\]

\State \textbf{Cosine Similarity Calculation}

Compute cosine similarities between the query vectors and context matrices:

\textbullet~ TF-IDF Cosine Similarity:
\[
\mathbf{s}_{\text{TF-IDF}} = \cos(\mathbf{q}_{\text{TF-IDF}}, \mathbf{M}_{\text{TF-IDF}})
\]

\textbullet~ Transformer Cosine Similarity:
\[
\mathbf{s}_{\text{Transformer}} = \cos(\mathbf{e}_q, \mathbf{E}_C)
\]

\State \textbf{Ensemble Similarity Calculation}

\textbullet~ Combine the similarity scores using predefined weights:
\[
\mathbf{s}_{\text{ensemble}} = 0.6 \cdot \mathbf{s}_{\text{TF-IDF}} + 0.4 \cdot \mathbf{s}_{\text{Transformer}}
\]

\State \textbf{Top-N Document Retrieval}

 \textbullet~ Retrieve the indices of the top \( n \) documents:
\[
\text{top}_n = \text{argsort}(\mathbf{s}_{\text{ensemble}})[-n:]
\]

 \textbullet~ Retrieve the corresponding contexts:
\[
\text{top}_n \text{ contexts} = [C_i \text{ for } i \text{ in top}_n]
\]

\end{algorithmic}
\end{algorithm}

In addition to the integration of sentence transformers, we employ NLP preprocessing techniques to further enhance the accuracy of the document retrieval process. These techniques include tokenization and stemming. Tokenization involves breaking down the text into individual words or tokens. Stemming reduces words to their base or root form, helping to normalize the text and handle variations of the same word. By applying these preprocessing steps, we aim to focus on the most informative terms and improve the precision of the retrieved documents.

\begin{algorithm}[!ht]
\caption{Ensemble Approach for Document Reader Module}
\begin{algorithmic}[1]

\State Initialize three models: Two BERT models (BERT1, BERT2) and one DeBERTa model

\vspace{0.5em}
\State \textbf{Training Phase:}
\vspace{0.5em}

\hspace{-0.6cm}\textbf{for} each model $m \in \{\text{BERT1}, \text{BERT2}, \text{DeBERTa}\}$ \textbf{do}
    
    \vspace{0.5em}
    \textbullet~ Fine-tune model $m$ on the question-answering task:
    \vspace{0.5em}
    
    \textbullet~ Generate hidden representations: $H_m = m(x)$, where $x = [x_1, x_2, \ldots, x_n]$ is the input sequence
    \vspace{0.5em}
    
    \textbullet~ Compute start and end probabilities for each token:
    \vspace{0.5em}
    
     \[
    \mathbf{P_{\text{start}}^{(m)}(i) = \text{softmax}(W_{\text{start}}^{(m)} \cdot h_i^{(m)})}\]
    
     \[
    \mathbf{P_{\text{end}}^{(m)}(i) = \text{softmax}(W_{\text{end}}^{(m)} \cdot h_i^{(m)})}\]
    \vspace{0.5em}
    
    \textbullet~ where $W_{\text{start}}^{(m)}$ and $W_{\text{end}}^{(m)}$ are learnable weight matrices
    \vspace{0.5em}
    
    \textbullet~ Update model parameters to minimize the loss function
    
\hspace{-0.6cm}\textbf{end for}

\vspace{1em}
\State \textbf{Testing and Evaluation Phase:}
\vspace{0.5em}

\hspace{-0.6cm} \textbf{for} each question in the dev and test set \textbf{do}

    \vspace{0.5em}
    \textbullet~ Generate predictions using each fine-tuned model:
    
    \[
    \mathbf{A_1 = \text{BERT1}(\text{question}, \text{context})}\]
    \[
    \mathbf{A_2 = \text{BERT2}(\text{question}, \text{context})}\]    
    \[
    \mathbf{A_3 = \text{DeBERTa}(\text{question}, \text{context})}\]
    
    \textbullet~ Compute ensemble prediction:
    
     \[
    \mathbf{A_{\text{ensemble}} = A_1 \cup A_2 \cup A_3}\]
    
    \textbullet~ Evaluate ensemble performance:
    \vspace{0.5em}
    
    \textbullet~ Compare $A_{\text{ensemble}}$ with ground truth answer
    \vspace{0.5em}
    
    \textbullet~ Categorize prediction as:
    \vspace{0.5em}
    
    \textbullet~ a. Correct: Exact match with ground truth
    \vspace{0.5em}
    
    \textbullet~ b. Similar: Significant overlap with ground truth
    \vspace{0.5em}
    
    \textbullet~ c. Incorrect: No match or significant overlap

\hspace{-0.6cm} \textbf{end for}

\vspace{0.5em}
\State Compute overall performance metrics (e.g., accuracy, F1 score)

\end{algorithmic}
\end{algorithm}

\subsection{Document Reader Module}

We propose an ensemble approach for the Document Reader module (depicted in Figure 2) that combines the strengths of multiple extractive question-answering models. Our ensemble consists of two BERT models with different initializations and one large DeBERTa model. BERT (Bidirectional Encoder Representations from Transformers) and DeBERTa (Decoding-enhanced BERT with Disentangled Attention) are transformer-based models that perform state-of-the-art natural language processing tasks, including question answering. The Algorithm 2 describes the workings of the BERT and DeBERTa models and then describe how we ensemble them for evaluating dev and test sets.

This algorithm presents the steps for an ensemble approach for the Document Reader module in a question-answering system. It combines two large BERT (Bidirectional Encoder Representations from Transformers) models and one large DeBERTa (Decoding-enhanced BERT with Disentangled Attention) model to leverage their strengths and potentially improve overall performance. The algorithm is divided into two main phases:

\vspace{0.5em}

\begin{enumerate}
    \item \textbf{Training Phase:} Each model (BERT1, BERT2, and DeBERTa) is fine-tuned separately on the question-answering task. It involves generating hidden representations, computing start and end probabilities for answer spans, and updating model parameters to minimize the loss function.
    \vspace{0.5em}
    
    \item \textbf{Testing and Evaluation Phase:} During inference, for each question in the test and validation set, predictions are generated using all three fine-tuned models. The ensemble prediction is formed by taking the union of predicted answers from all models. The algorithm then evaluates the ensemble performance by comparing predictions with ground truth answers and categorizing them as correct, similar, or incorrect.
\end{enumerate}

\vspace{0.5em}

This ensemble method aims to enhance the accuracy and reliability of the question-answering system by combining multiple state-of-the-art models. The diversity in predictions can lead to more robust results, potentially capturing nuances that individual models might miss.

\vspace{1em}

\section{Experimental setup}

We evaluate the performance of our proposed model by utilizing the BenthamQA and HW-SQuAD datasets. By combining the OCR texts, we have created the context for our recognition-based pipeline. The question and the answer are derived from these two datasets. Following the extraction of context, questions, and answers, we implemented the basic preprocessing approach. Afterwards, we implemented our proposed Document Retriever. In order to implement the Recognition-based model, we converted the output to SQuAD dataset format. The HW-SQuAD dataset contains over 84,000+ pairs of QA pairs, while the Bentham-QA dataset contains 200 pairs. For training, validation, and testing, we have implemented the exact same split ratio as in \cite{HW-SQUaD}. WE have fintuned our model on HW-SQuAD dataset. The BenthamQA dataset remains unchanged and only used for evaluation.

Our training was conducted using four 2080 ti Nvidia RTX graphics cards. AdamW is employed for BERT large, while Adam Optimizer is employed for the DeBERTa model. 

\subsection{Evaluation Metrics}

We utilized the top 5-accuracy scores for evaluating the Document Retriever Module, as it is implemented in \cite{HW-SQUaD}. Top 5 accuracy is calculated as:

\begin{equation}
Top\_5\_Accuracy = \frac{1}{N} \sum_{i=1}^{N} I(ground\_truth_i \in top\_5\_predictions_i)
\end{equation}

where:
\begin{itemize}
\item $N$: Total number of instances
\item $I$: Indicator function that equals 1 if the ground truth is present in the top 5 predictions, and 0 otherwise
\end{itemize}

We employed two evaluation metrics for Document Reader: the F1 score and the EM \cite{rajpurkar-etal-2016-squad}\cite{HW-SQUaD}. The F1 score is the harmonic mean of precision and recall:

\begin{equation}
F1 = 2 \times \frac{precision \times recall}{precision + recall}
\end{equation}

where:
\begin{itemize}
\item Precision = $\frac{TP}{TP + FP}$
\item Recall = $\frac{TP}{TP + FN}$
\item TP: True Positives
\item FP: False Positives
\item FN: False Negatives
\end{itemize}

The another metric, Exact Match (EM) is calculated as:

\begin{equation}
EM = \frac{1}{N} \sum_{i=1}^{N} I(predicted_i = ground\_truth_i)
\end{equation}

where:
\begin{itemize}
\item $N$: Total number of instances
\item $I$: Indicator function that equals 1 if the predicted answer matches the ground truth, and 0 otherwise
\end{itemize}

\section{Results and Analysis}

This section presents a three-part analysis of our experimental results. First, we investigate the effect of TF-IDF and sentence transformers on document retrieval performance. Second, we evaluate the proposed Document Reader's performance on the HW-SQuAD and BenthamQA datasets. Finally, we conduct ablation studies to examine the impact of critical components on overall QA performance and provide an error analysis to identify common errors and propose mitigation strategies. This section offers insights into the effectiveness of our approach and its performance on benchmark datasets, contributing to a better understanding of question answering on handwritten documents.

\begin{figure*}[!ht]
\hspace{-1.9cm}
    \includegraphics[width=1.25\textwidth]{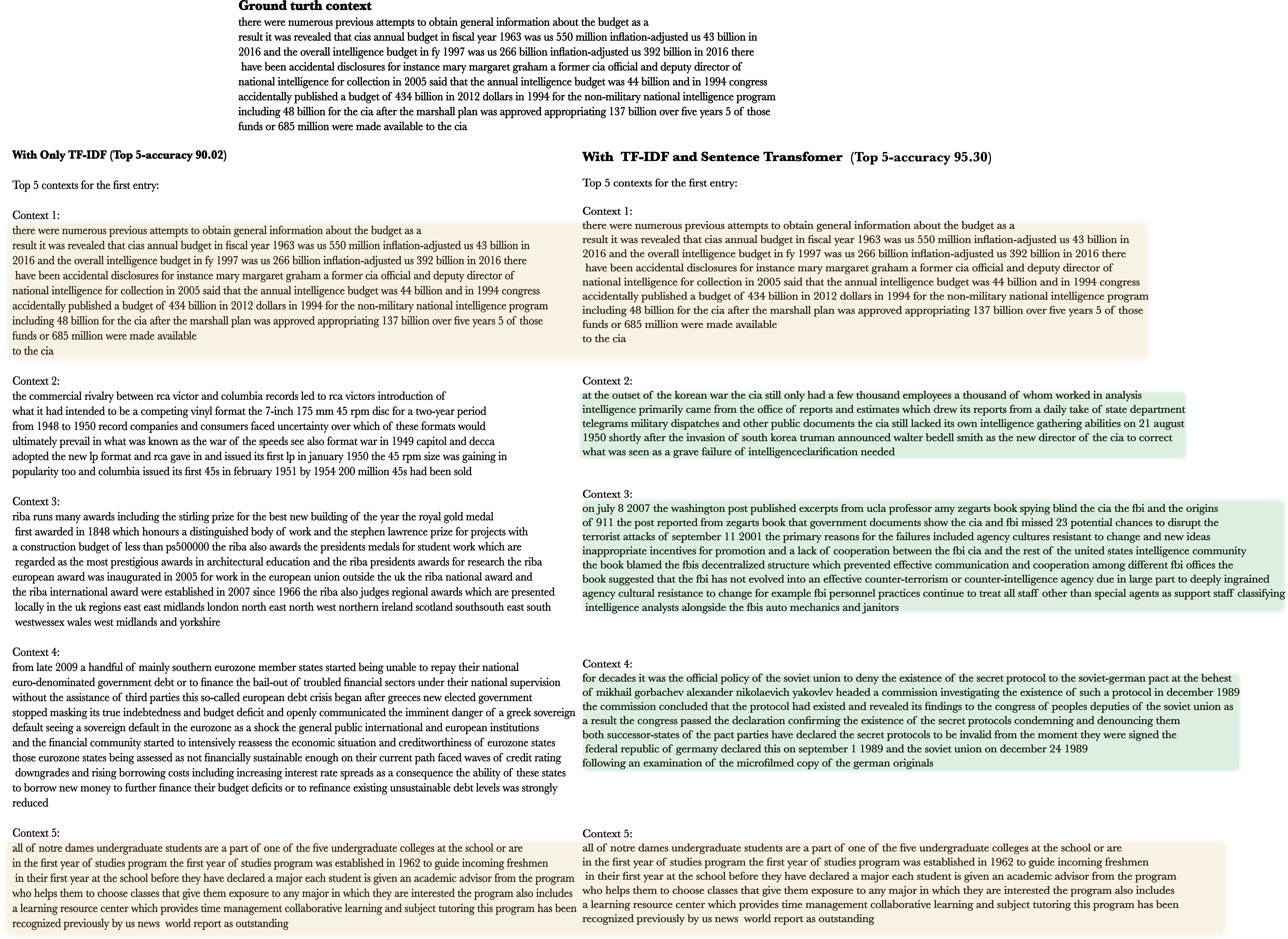}
    \caption{Comparing two approaches of Document retrieval. The common passage (from HW-SQuAD) for both are highlighted with light brown and the improved retrieved passages are highlighted by light green.}
    \label{fig:enter-label}
\end{figure*}

\subsection{Impact of Sentence Transformers and NLP Preprocessing on Document Retrieval}

In this subsection, we discuss the effect of incorporating sentence transformers and NLP preprocessing techniques alongside the TF-IDF algorithm for document retrieval. The previous model's document retrieval component relied solely on TF-IDF, achieving a top-5 accuracy of approximately 90\%. By integrating NLP preprocessing and sentence transformers, we aim to enhance the retrieval performance. Table 1 presents the results obtained on the HW-SQuAD and BenthamQA datasets after applying these modifications.

\begin{figure*}[ht]
\hspace{-1.5cm}
    \includegraphics[width=1.2\textwidth]{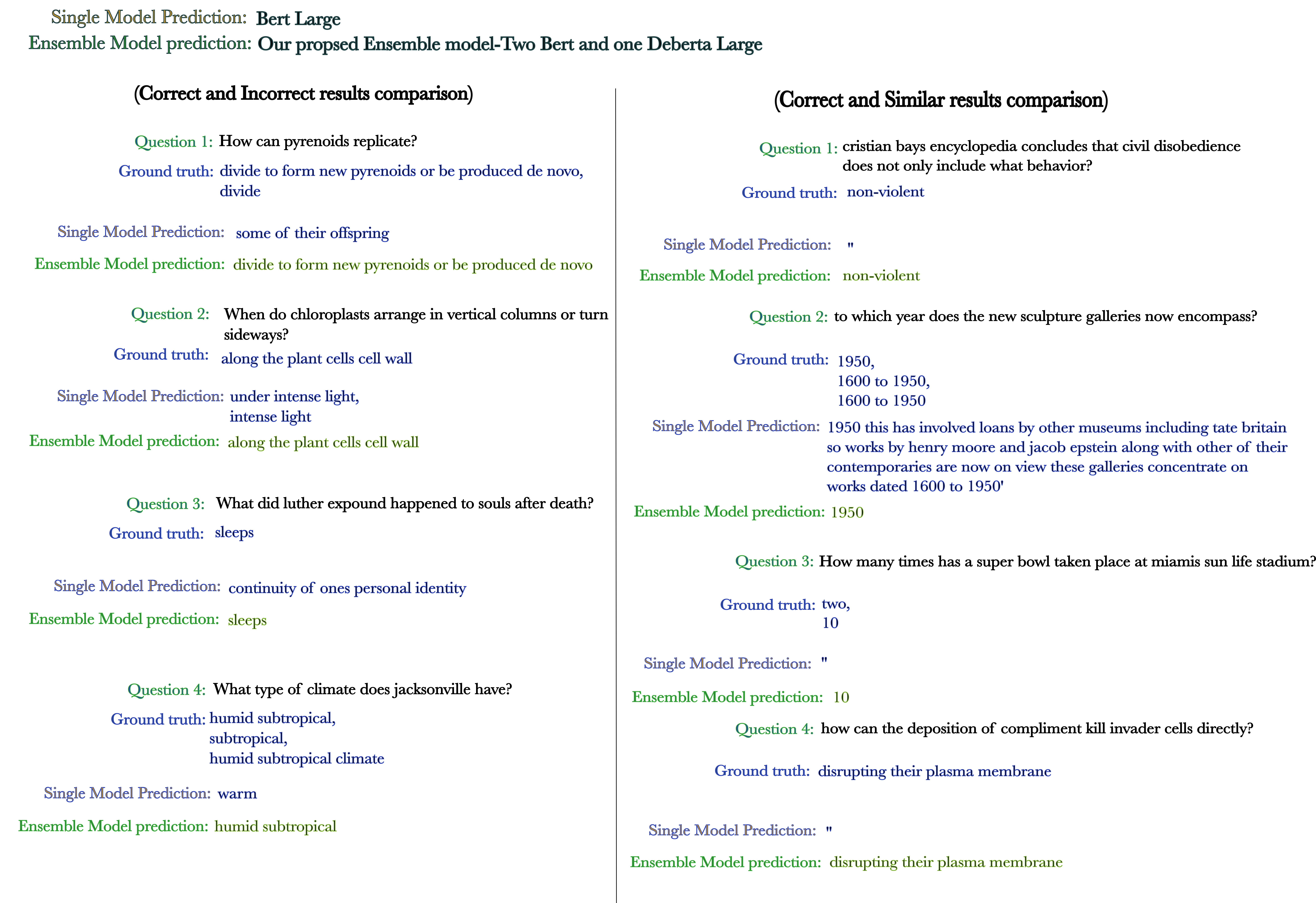}
    \caption{Comparison predicted result between the single and ensemble models (from HW-SQuAD). Left side depicts the scenario where the single model gives and the ensemble gives the correct result. The right side depicts the correct and similar case. }
    \label{fig:enter-label}
\end{figure*}

\begin{figure*}[ht]
\hspace{-1.5cm}
    \includegraphics[width=1.2\textwidth]{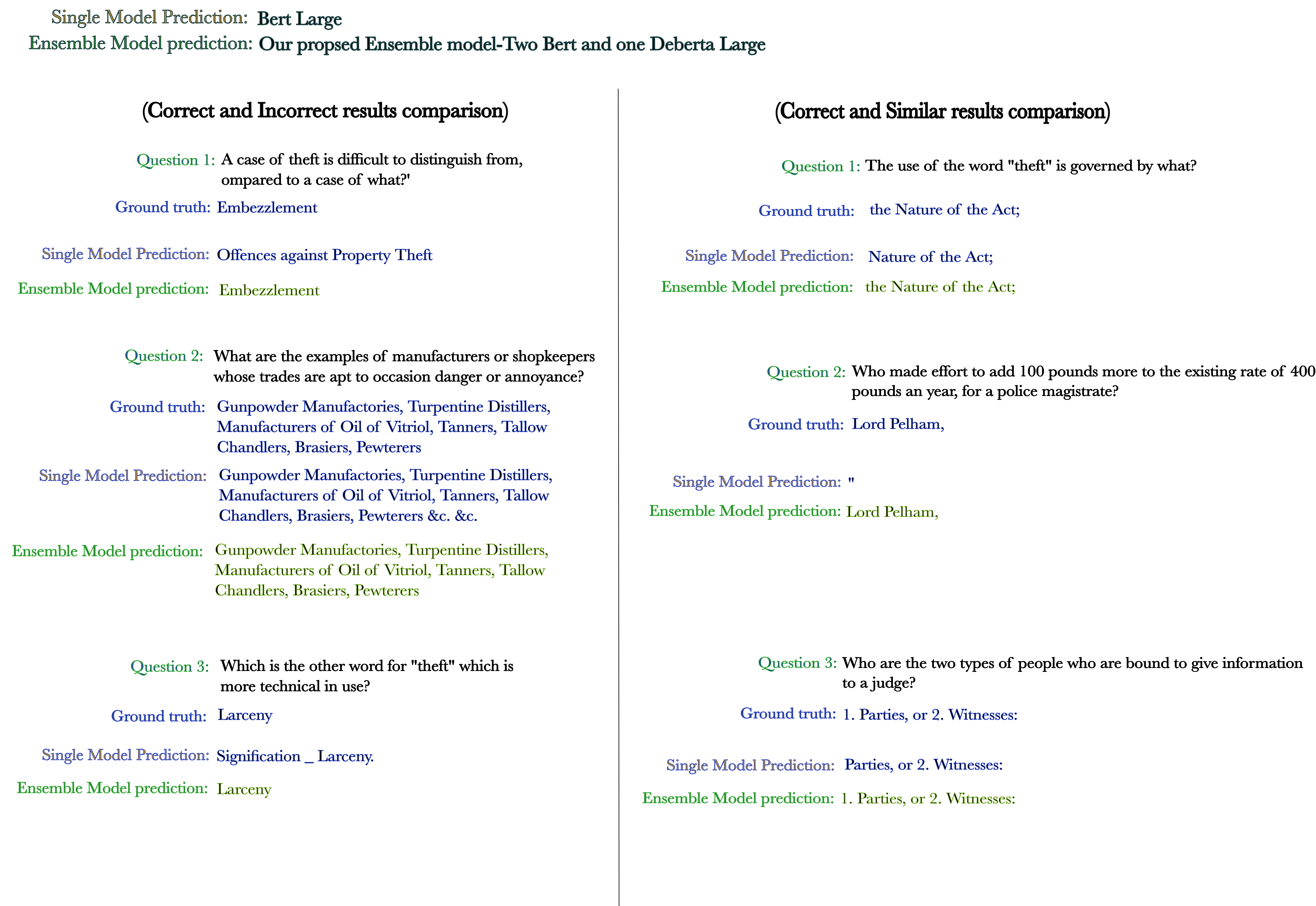}
    \caption{Comparison predicted result between the single and ensemble models (from BenthamQA). Left side depicts the scenario where the single model gives and the ensemble gives the correct result. The right side depicts the correct and similar case. }
    \label{fig:enter-label}
\end{figure*}

The results demonstrate that the inclusion of sentence transformers and NLP preprocessing techniques leads to a significant improvement in document retrieval accuracy. On the HW-SQuAD dataset, our approach achieves a top-5 accuracy of 95.3\%, surpassing the previous model's performance by 4.83\%. Similarly, on the BenthamQA dataset, we observe an increase in top-5 accuracy from 98.5\% to 99.5\%. These findings highlight the effectiveness of leveraging sentence transformers and NLP preprocessing in capturing semantic similarities between the question and the document, enabling more precise retrieval of relevant documents.

\subsubsection{Comparison of the two retrieval on real-world examples}

In Fig. 2 we have compared the two technique side by side on an example. We take one ground truth context (depicted on the top of the Figure) and corresponding top 5 contexts retrieved by two model. The common contexts are highlighted as light orange and the improved retrived contexts are highlighted with light green. Though Both the combined approach (TF-IDF, sentence transformer) and the TF-IDF-only method successfully retrieved the ground truth context as their top result the old approach lacks in semantic relevance, thematic consistency. We give the detail analysis in the below sections.

\textit{Semantic relevance of the retrieved contexts:} The combined approach appears to retrieve contexts that are more semantically related to the ground truth context and the general theme of the CIA and intelligence agencies. For example, the second context discusses the CIA's lack of intelligence-gathering abilities during the Korean War, while the third context mentions the CIA and FBI's missed opportunities to prevent the 9/11 attacks. These contexts, although not directly related to the CIA's budget, are still relevant to the overall topic of the CIA and its performance.

\textit{Thematic consistency of the retrieved contexts:} The combined approach demonstrates better thematic consistency among the retrieved contexts. Apart from the fifth context (which appears to be an outlier), the other contexts are related to the CIA, intelligence agencies, or historical events involving them. This suggests that the combined approach is more effective at capturing the overall theme of the query.

\textit{Reduced Reliance on Exact Keyword Matching:} The combined approach reduces the reliance on exact keyword matching by leveraging sentence transformers and count vectors. This allows for the retrieval of relevant contexts that may use synonyms, paraphrases, or related terms instead of the exact keywords present in the query. By capturing the semantic similarity between words and phrases, the combined approach can identify relevant contexts that would be missed by a purely keyword-based method like TF-IDF.

In contrast, The TF-IDF-only method has several disadvantages compared to the combined approach. It relies solely on keyword matching, lacking the ability to capture semantic meaning and contextual relationships between words. As a result, the retrieved contexts may be less relevant to the query, as seen in the examples of the commercial rivalry between ``RCA Victor'' and ``Columbia Records'' or the ``European debt'' crisis, which are unrelated to the ``CIA'' or its budget. The TF-IDF-only method also shows less thematic consistency, retrieving contexts spanning various disconnected topics based on keyword overlap alone. This lack of coherence can lead to a less focused and relevant set of results. Furthermore, the method is sensitive to keyword variations and may struggle to retrieve relevant contexts that use synonyms or related terms, potentially omitting valuable information.

In conclusion, the combined approach, which incorporates TF-IDF, sentence transformers, and count vectors, offers significant advantages over the TF-IDF-only method. By capturing semantic meaning, thematic consistency, and reducing the reliance on exact keyword matching, the combined approach retrieves more relevant and coherent contexts. On the other hand, the TF-IDF-only method's limitations, such as limited semantic understanding, lack of thematic consistency, and sensitivity to keyword variations, can lead to the retrieval of irrelevant or disconnected contexts. Therefore, the combined approach demonstrates superior performance in identifying relevant and meaningful contexts for a given query, providing a more comprehensive and accurate representation of the information sought.

\begin{table}[h]
    \centering
    \caption{Results of our document retriever on transcriptions of
the documents in HW-SQuAD and BenthamQA}
    \resizebox{0.7\textwidth}{!}{%
        \begin{tabular}{lccc}\toprule
    \multirow{2}{*}{Transcriptions} & \multicolumn{2}{c}{Test} &\\
    & HW-SQuAD & BenthamQA \\ 
    \midrule
    a. TF-IDF \cite{HW-SQUaD} & 90.2 & 98.5  \\
    b. TF-IDF + ST + Preprocessing (proposed) & 95.30 & 100 \\
    
    \bottomrule
    \end{tabular}%
    }
\end{table}

\begin{table}[h]
    \centering
    \caption{Result of applying Our proposed models on HW-SQuAD and BenthamQA}
    \resizebox{\textwidth}{!}{%
    \begin{tabular}{lccccc}\toprule
    \multirow{2}{*}{Model} & \multicolumn{2}{c}{HW-SQuAD } & {BenthamQA}\\
    & F1 & Exact Match & F1 & Exact Match \\ 
    \midrule
    a. TF-IDF + BERT \cite{HW-SQUaD} & 76.82 & 70.73  & 78.41 & 66.00\\
    b. TF-IDF + ST + preprocessing + Ensemble (proposed) & 90.10 & 82.02 & 81.57 & 69\\
    \bottomrule
    \end{tabular}%
    }
\end{table}

\subsection{Impact of applying Ensemble method in Document Reader} 

We present a comparative performance analysis between the baseline model referenced in \cite{HW-SQUaD} and our proposed model, evaluating their efficacy on both the HW-SQuAD and BenthamQA datasets. The results of this comparison are summarized in Table 2. Performance metrics utilized in this assessment include the F1 score and Exact Match (EM) percentage on the respective test sets.

Our proposed model demonstrates superior performance across both datasets. On the HW-SQuAD dataset, it achieves an F1 score of 90.10\% and an Exact Match of 82.02\%, representing a significant improvement over the baseline model's 76.82\% F1 and 70.73\% EM. Similarly, on the BenthamQA dataset, our model attains a 81.57\% F1 score and 69\% EM, substantially outperforming the baseline's 78.41\% F1 and 66\% EM.

The enhanced performance of our proposed models can be attributed to Sentence Transformer (ST) and an ensemble approach, building upon the standard TF-IDF + BERT baseline for question answering tasks on handwritten document datasets. The ensemble model's exceptional results, approaching 90\% F1 and exceeding 80\% Exact Match, establish a new benchmark for performance on these challenging datasets. The methodologies introduced in this study have the potential to advance state-of-the-art question-answering systems for handwritten documents.

\subsection{Analysis of predicted results of our proposed model}

 In this section, we analyze the effect of applying the ensemble method to our proposed model. We have taken a few questions from the HW-SQuAD and BenthamQA datasets and compared the results in Fig. 5 and 6. respectively.
In the HW-SQuAD dataset, the ensemble model consistently corrects predictions where the single model falters. For instance, in responses requiring specific information such as ``divide to form new pyrenoids or be produced de novo,'' the single model inaccurately predicts ``some of their offspring.'' In contrast, the ensemble model accurately matches the ground truth. This precision is also evident in questions requiring detailed understanding, such as predicting ``along the plant cells cell wall'' instead of ``under intense light.'' Additionally, for broader categorical answers like ``humid subtropical,'' the ensemble model refines the single model's vague prediction of ``warm'' to the precise ground truth. This meticulous correction and alignment with the expected answers underline the ensemble model's enhanced interpretative capabilities.
The effectiveness of the ensemble model is equally prominent in the BenthamQA dataset. The ensemble model not only rectifies broader, less accurate predictions of the single model but also ensures precision in complex queries. For example, it refines the single model's generic prediction of ``Offences against Property Theft'' to the specific term ``Embezzlement,'' demonstrating its ability to grasp nuanced legal terminology. Moreover, for questions requiring multiple valid responses, such as listing various historical manufacturers or recognizing titles like ``Lord Pelham,'' the ensemble model accurately captures all relevant details, showcasing its comprehensive understanding and reliability.

Overall, the ensemble model's proficiency in reducing incorrect and similar incorrect results significantly enhances its reliability and precision. By amalgamating the strengths of multiple models, it ensures that predictions are more accurate and consistent with the ground truth. This integration allows the ensemble model to capture a broader spectrum of linguistic and contextual nuances, ultimately leading to more robust and dependable AI systems. 

\begin{figure*}[!ht]
\hspace{-0.6cm}
    \includegraphics[width=\textwidth]{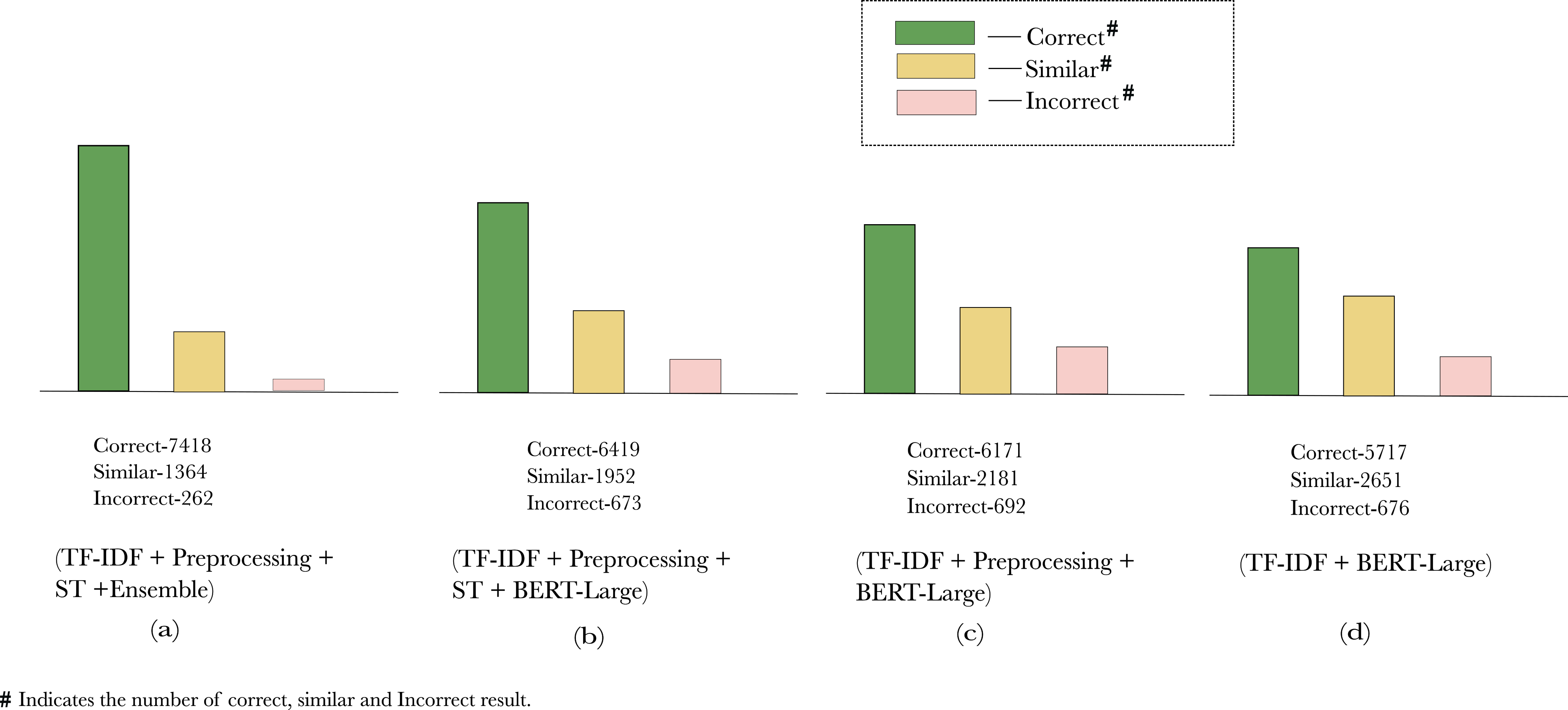}
    \caption{Comparison of correct, similar and incorrect among all the models. The addition of our proposed methodology improves the no of corrects and reducing the no of similar and incorrect result. We have depicted the how addition of each technique over the old model improves the performance}
    \label{fig:enter-label}
\end{figure*}

\begin{table}[h]
    \centering
    \caption{Ablation studies of Our Document Retriever Module.}
    \resizebox{0.6\textwidth}{!}{%
    \begin{tabular}{lccc}\toprule
    \multirow{2}{*}{Model Details} & \multicolumn{2}{c}{Top-5 accuracies} &\\
    & HW-SQuAD & BenthamQA \\ 
    \midrule
     a. TF-IDF  & 90.2 & 98.5  \\
     b. TF-IDF + pre-processing & 90.8 & 98.94   \\
     c. TF-IDF + pre-processing + ST & 95.30 & 100 \\
    
    \bottomrule
    \end{tabular}%
    }
\end{table}

\begin{table}[h]
    \centering
    \caption{Ablation studies of Our Document Reader Module on HW-SQuAD}
    \resizebox{0.7\textwidth}{!}{%
    \begin{tabular}{lccc}\toprule
    \multirow{2}{*}{Model Details} & \multicolumn{2}{c}{HW-SQuAD} &\\
    & F1 & Exact Match \\ 
    \midrule
    a. TF-IDF + BERT \cite{HW-SQUaD} & 76.82 & 70.2  \\
    b. TF-IDF + BERT (Our implementation) & 77.46 & 63.21  \\
    c. TF-IDF + preprocessing + BERT (proposed) & 81.18 & 68.35 \\
    d. TF-IDF + preprocessing + ST + BERT (proposed) & 83.20 & 71.33   \\
     e. TF-IDF + preprocessing + ST + Ensemble (proposed) & 90.10 & 82.02 \\
    
    \bottomrule
    \end{tabular}%
    }
\end{table}

\begin{table}[h]
    \centering
    \caption{Ablation studies of Our Document Reader Module on BenthamQA}
    \resizebox{0.7\textwidth}{!}{%
    \begin{tabular}{lcc}
    \toprule
    \multirow{2}{*}{Model Details} & \multicolumn{2}{c}{BenthamQA} \\
    & F1 & Exact Match \\ 
    \midrule
    a. TF-IDF + BERT \cite{HW-SQUaD} & 78.41 & 66 \\
    b. TF-IDF + BERT (Our implementation no finetuning) & 64.75 & 53 \\
    c. TF-IDF + preprocessing + BERT (proposed) & 73.96 & 59.2\\
    d. TF-IDF + ST + preprocessing + BERT (proposed) & 73.96 & 59.2 \\
    e. TF-IDF + ST + preprocessing + ensemble (proposed) & 81.57 & 69.1 \\
    \bottomrule
    \end{tabular}%
    }
\end{table}

\subsection{Ablation Studies}

Tables 3 and 4 illustrate the ablation studies that were conducted on the HW-SQuAD and BenthamQA datasets to assess the efficacy of various components in our proposed model.

As an initial step, we investigated the impact of preprocessing and the Sentence Transformer on our Document Retriever Module. Using only the TF-IDF and basic preprocessing, the baseline model \cite{HW-SQUaD} achieved 90.2 top-5 accuracy. We have implemented advanced preprocessing techniques, resulting in a top-5 accuracy score of 90.8. Furthermore, we implemented the Sentence Transformer (ST), which has a significant effect on performance, resulting in a 5\% increase. Additionally, in the BenthamQA dataset, improving the preprocessing results in an approximate 0.5\% increase in top-5 accuracy, and when the ST is added to top of that, it reaches 100\%.

Next we have done ablation studies for the Document Reader part for both dataset HW-SQuAD and BenthamQA. Fird we describe the effect on the HW-SQuAD dataset which is represented in Table 4. Here we have listed the result reported in \cite{HW-SQUaD} and the we have given our results.
At first we have applied the single BERT model and we reached 63.21 Exact Match and 77.46 F1 score without any type of preprocessing. when we add our proposed preprocessing to the Retrieval pipeline the it reached 68.21 EM and 77.46 F1 score.
By adding the Sentence Transformer (ST) module to the retrieval pipeline, the model gets improved and achieved F1 score of 83.20 and the Exact Match of 71.33. This demonstrates the effectiveness of incorporating semantic information for retrieving more relevant documents.

The ablation studies in the BenthamQA dataset gets presented in the Table 5. Like the Table 4 we first give the result in the HW-SQuAD paper. Next we have given the result of our implementation. Then we have add our proposed techniques and how it improves the performance. When we slightly finetuned and add the preprocessing the model scored 82.97 EM. When we add the ST to it the EM improves further to 86.17\%. Like in HW-SQuAD Semantic information in time of retrieval improves the performance of Document reader significantly.

Next, we analyzed the performance of our ensemble approach, which includes three extractive QA models (two BERT and one DeBERTa large) in addition to the retrieval enhancements. The TF-IDF + ST + Ensemble model achieved an impressive F1 score of 90.10 and an Exact Match of 82.02, significantly outperforming the baseline. Also in BenthamQA dataset, when we add the ensemble method we reach over 90\% EM and 96.12\% F1 score. These results highlight the importance of both the improved document retrieval and the ensemble strategy in our proposed approach.

\subsubsection{Analysis of Correct, Similar, and Incorrect Matches
}

In addition to the overall results, we have analyzed the number of correct, similar, and incorrect matches for each case mentioned in Table 3. Figure 3 provides a visual depiction of these results. Our proposed model yields 7,418 correct matches, 1,364 similar matches, and 262 incorrect matches, representing the lowest number of similar and incorrect matches among all the models evaluated. In comparison, the old model produces 676 incorrect matches, nearly three times the number generated by our proposed approach.

The incorporation of a Sentence Transformer in the Document Retriever stage leads to an increase in the number of correct matches to 6,419, an improvement of over 700 compared to the old model. Concurrently, the number of incorrect matches is reduced to 1,952. The application of an ensemble technique further decreases the number of incorrect matches to 1,364 while increasing the number of correct matches to 7,419.

Based on these observations, we can conclude that adding a Sentence Transformer in the Document Retriever stage significantly enhances the quality of the retrieved documents. Consequently, even a single BERT large model remarkably improves the Document Reader component. The implementation of an ensemble method further improves the Document Reader's semantic understanding of the context, resulting in a three-fold decrease in the number of incorrect matches compared to the old model and a reduction by half in Similar matches.

Overall, the ablation studies validate the efficacy of the key components in our model. The combination of semantic similarity-enhanced retrieval and an ensemble of strong reader models leads to state-of-the-art performance on the HW-SQuAD benchmark.

\section{Conclusion}

This paper proposes a novel approach for answering questions on handwritten documents by combining advanced document retrieval techniques with an ensemble of extractive QA models. Our enhanced document retriever, which leverages the strengths of TF-IDF and sentence transformers, significantly improves retrieval performance. The ensemble-based document reader, utilizing BERT and DeBERTa large models, enables robust and accurate answer extraction from the retrieved documents.

Experimental results on the HW-SQuAD and Bentham-QA datasets demonstrate the effectiveness of our approach. They surpass previous recognition-based methods and set a new benchmark for QA performance on handwritten documents. However, room remains for further improvement, such as exploring recognition-free methods and techniques to handle noise and varying writing styles.

Our work is a significant step forward in the field of question-answering on handwritten documents. By presenting a novel approach and achieving superior results, we have not only advanced the state of the art but also opened up new avenues for research in this challenging and vital area of natural language processing.

\clearpage

\bibliographystyle{IEEEtran}
\bibliography{egbib}

\end{document}